# SANet: Selective Attention Network for Infrared Small Target Detection

Yingmei Zhang, Wangtao Bao, Qin Xiao, Yong Yang, *Senior Member, IEEE*, Weiguo Wan, *Member, IEEE*, Yitao Luo, Xueting Zou, and Lei Zhang

***Abstract*—Infrared small target detection aims to accurately identify and locate dim targets in complex backgrounds and supports applications such as maritime surveillance and military search and rescue. However, the small size and weak contrast of infrared targets make it difficult to balance detection accuracy and false alarms. This paper proposes a selective attention network (SANet) for infrared small target detection. A dual-path semantic-aware module combines standard and pinwheel-shaped convolutions to preserve local spatial consistency and capture broader contextual information. Spatial and channel attention further refine the features and improve target-background discrimination. To address the limitations of static skip connections in U-Net, a selective attention fusion module adaptively integrates features across scales using spatially varying weights. It selectively enhances salient regions and improves discrimination between true targets and false alarms. Experiments on three public benchmarks, NUAA-SIRST, IRSTD-1K, and NUDT-SIRST, show that SANet achieves competitive performance in intersection over union (IoU), normalized IoU, detection probability, and false alarm rate. Its IoU exceeds that of the second-best method by 1.93, 4.32, and 2.21 percentage points, respectively. These results support the effectiveness of SANet in dim-target perception, discriminative feature representation, and background suppression. The source code is available at https://anonymous.4open.science/r/SANet-E808/ and https://gitcode.com/m0_61988291/SANet.**

***Index Terms*—Infrared small target detection, dual-path semantic-aware module, pinwheel-shaped convolution, selective attention fusion, spatially varying weights.**

## I. Introduction

Infrared small target detection (IRSTD) aims to accurately identify small, low-contrast targets in complex backgrounds [1]–[3]. It has broad practical value [4] and supports critical applications, including maritime surveillance, search and rescue, and early warning. In these applications, detection accuracy directly affects the speed and reliability of a system's response to potential threats and, consequently, mission outcomes and personnel safety. Improving the accuracy and robustness of IRSTD in complex scenes remains a central objective for practical deployment.

Traditional methods generally treat infrared small target detection as a filtering and target-enhancement problem, relying on handcrafted feature extraction and carefully tuned parameters. Their flexibility and robustness are often limited when target size, shape, and background interference vary substantially in real scenes. In recent years, deep learning techniques, including convolutional neural networks (CNNs), have substantially improved IRSTD accuracy. Dai et al. [5] propose a segmentation-based detector with asymmetric contextual modulation for feature fusion in a U-Net framework [6]. Li et al. [7] develop a dense nested attention network for multiscale feature fusion. Zhang et al. [8] introduce an infrared shape network that performs well on multiple datasets. Yang et al. [9] propose a progressive background-aware transformer that refines background context across multiple stages to mitigate feature degradation and error accumulation, improving detection and suppressing clutter. Wu et al. [10] replace conventional cross-layer feature fusion in U-Net with learnable local saliency kernels to guide the extraction of salient infrared target features.

Despite these advances, information bottlenecks in target feature extraction continue to limit detection performance. Many existing designs address individual deficiencies, while the underlying difficulties of infrared small target detection remain. Two limitations in network architecture and feature modeling are particularly relevant.

First, early convolutional layers in existing architectures often lack sensitivity to fine-grained information from very small targets [11]. This reduces feature contrast between targets and complex backgrounds, limiting the discrimination of weak targets and reducing detection accuracy and reliability.

Second, the static feature connections widely used in current networks lack the ability to adjust feature transfer dynamically in salient regions. Fixed fusion rules make it difficult to adaptively distinguish true infrared targets from regions that generate false alarms, restricting performance in complex backgrounds.

To address these limitations, we propose a selective attention network for infrared small target detection, termed SANet. We first develop a dual-path semantic-aware module (DSM) as the basic feature extraction unit to improve sensitivity to weak targets. DSM combines standard and pinwheel-shaped convolutions in two parallel paths. The standard convolution path preserves basic discriminative

This work was supported in part by the National Natural Science Foundation of China under Grants 62072218 and 62261025 and in part by the Early-Career Young Science and Technology Talent Training Project of Jiangxi Province under Grants 20244BCE52075 and 20244BCE52088. (Corresponding author: Qin Xiao.)

Y. Zhang, W. Bao, W. Wan, Y. Luo, X. Zou, and L. Zhang are with the School of Software and Internet of Things Engineering, Jiangxi University of Finance and Economics, Nanchang 330032, China (e-mail: zhangyingmei@jxufe.edu.cn; baowangtao29@gmail.com; wanweiguo@jxufe.edu.cn; 3014815803@qq.com; zouxueting0113@126.com; 13647037937@163.com).

Q. Xiao is with the School of Information Management and Mathematics, Jiangxi University of Finance and Economics, Nanchang 330032, China (e-mail: xiaoqin@jxufe.edu.cn).

Y. Yang is with the School of Computer Science and Technology, Tiangong University, Tianjin 300387, China (e-mail: yangyong@tiangong.edu.cn).

features, whereas the pinwheel-shaped convolution path [12] expands the receptive field to enhance responses to target regions. The fused features are then refined by the spatial and channel attention mechanisms of the convolutional block attention module (CBAM) [13], which emphasize useful information and suppress background noise.

Distinguishing true targets from false alarms requires an analysis of the surrounding background beyond local image regions. We therefore introduce a selective attention fusion module (SAFM) to replace static skip connections in conventional U-Net. SAFM combines selective attention with a broad receptive field and dynamically integrates shallow and deep features according to the context. This improves discrimination between true small targets and background interference in complex scenes. The main contributions are as follows.

1) We propose an infrared small target detection method based on selective attention. It improves the distinction between true targets and false detections in complex backgrounds, increasing detection accuracy and robustness.

2) We construct DSM by combining standard and pinwheel-shaped convolutions. Its broader receptive field helps suppress background noise, while spatial and channel attention improve the representation of fine-grained features.

3) We design SAFM to replace static skip connections in conventional U-Net. Dynamic feature fusion selectively guides information transfer and improves false alarm suppression.

4) Extensive experiments on three public datasets validate the proposed method. SANet compares favorably with existing state-of-the-art detectors across multiple performance metrics.

## II. Related Work

### A. Traditional Methods for IRSTD

Traditional methods form the foundation of research on infrared small target detection. Early approaches rely primarily on handcrafted image processing and statistical models and offer some robustness under low-resolution and low-signal-to-noise conditions. Filter-based methods [14] suppress background interference through spatial- or frequency-domain filtering. They perform well on relatively uniform backgrounds, but their effectiveness is limited when target and background frequency components overlap. Low-rank methods [15], [16] decompose an image into a low-rank background and a sparse target component, enhancing targets at low signal-to-noise ratios. However, separating sparse noise from true targets remains difficult and can lead to high false alarm rates. Local contrast methods [17], [18] enhance target saliency by measuring intensity differences in local regions. Although effective for small targets, their reliance on fixed window sizes limits adaptation to targets at different scales.

Most traditional approaches depend on fixed handcrafted features and parameter settings. They cannot fully model the diversity and uncertainty of targets in complex backgrounds, which limits their accuracy and robustness in practical applications. These limitations motivate data-driven methods.

### B. Deep Learning Methods for IRSTD

Compared with traditional approaches, data-driven deep learning has substantially advanced infrared small target detection. MDvsFA [19] introduces adversarial learning to alleviate the trade-off between missed detections and false alarms. ACM [5] uses asymmetric contextual modulation within U-Net to combine high-level semantics with fine-grained low-level details. ALCNet [20] embeds a parameter-free local contrast measurement module in the network to overcome the limited receptive field of small convolutional kernels. UIUNet [21] embeds small U-Net structures in a larger U-Net to learn multilevel and multiscale target representations. GCI-Net [22] uses Gaussian curvature information to enhance the geometric features of target regions. ABC [23] combines convolutional and linear structures with transformer modules to improve feature extraction and fusion while suppressing background noise. SCTransNet [24] uses a spatial-channel cross transformer to increase target-clutter separability across semantic levels. CFD-Net [25] draws on computational fluid dynamics, modeling infrared feature evolution as the flow of pixel particles to improve target representation. HDNet [26] combines multiscale target perception in the spatial domain with low-frequency suppression in the frequency domain to improve accuracy and robustness.

HAFNet [27] combines dual-branch semantic perception with hierarchical feature fusion in the encoder and decoder. MPCNet [28] integrates multiscale perception with cross-attention feature fusion. FSGNet [29] uses frequency-aware skip connections and global semantic guidance to suppress background interference and preserve semantic consistency. APTNet [30] combines double residual attention and adaptive partial transformer blocks to strengthen target representation and context modeling.

General-purpose segmentation models have also been introduced into IRSTD. IRSAM [31] adapts the Segment Anything Model [32] to infrared small target detection and incorporates a wavelet-based Perona–Malik diffusion module to enhance edge feature extraction. State space models (SSMs) [33] offer effective long-sequence modeling and provide an alternative for capturing long-range dependencies in vision tasks. Mamba [34], a representative SSM architecture, has attracted considerable attention. MiM-ISTD [35] is the first work to apply Mamba to infrared small target detection. Building on this direction, IRMamba [36] models pixel differences in intensity and direction between scanning positions and their central neighborhoods. These differences are incorporated into the state update equations to strengthen the representation of target details.

### A. Overall Network Architecture

Fig. 1 illustrates the overall architecture of SANet. The network is built on the conventional U-Net framework and incorporates two key components, DSM and SAFM, to improve detection performance.

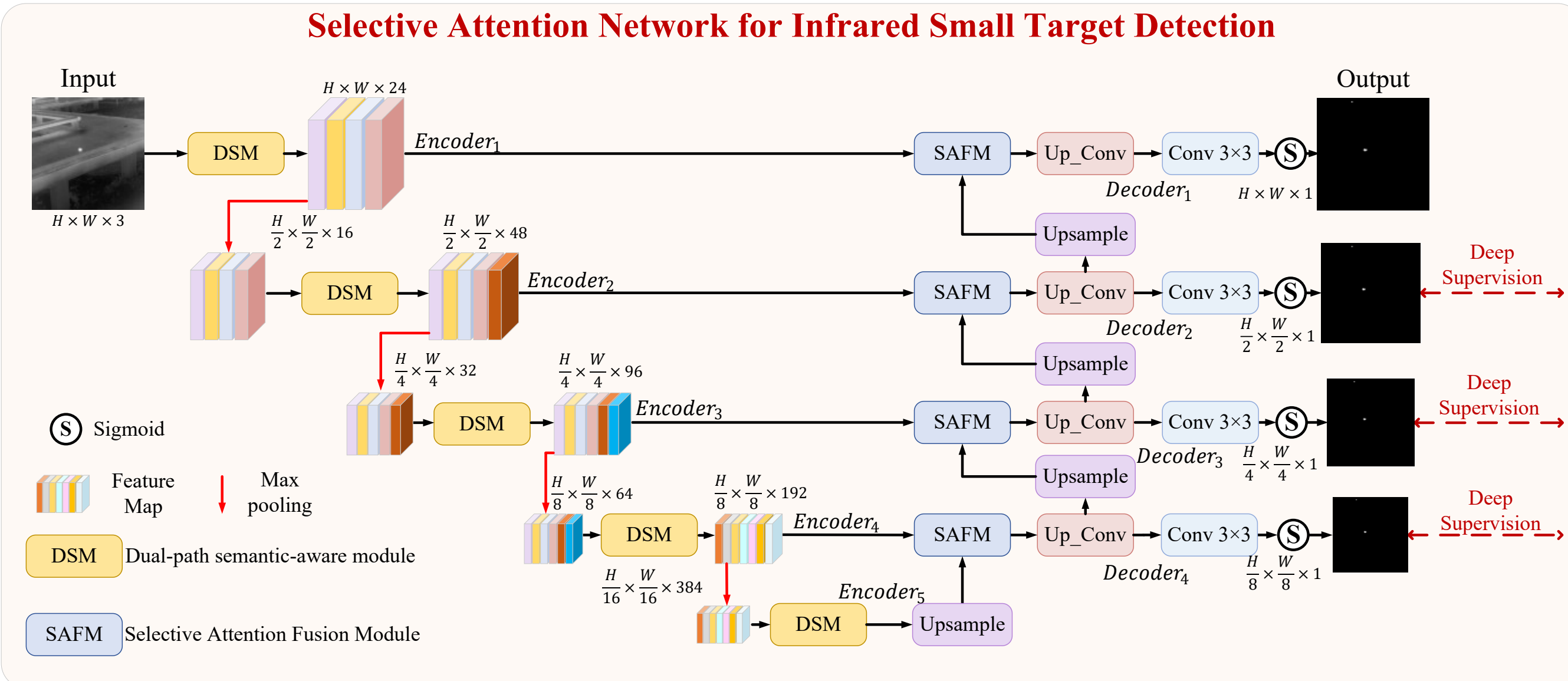

Fig. 1. Overall Framework of the SANet.

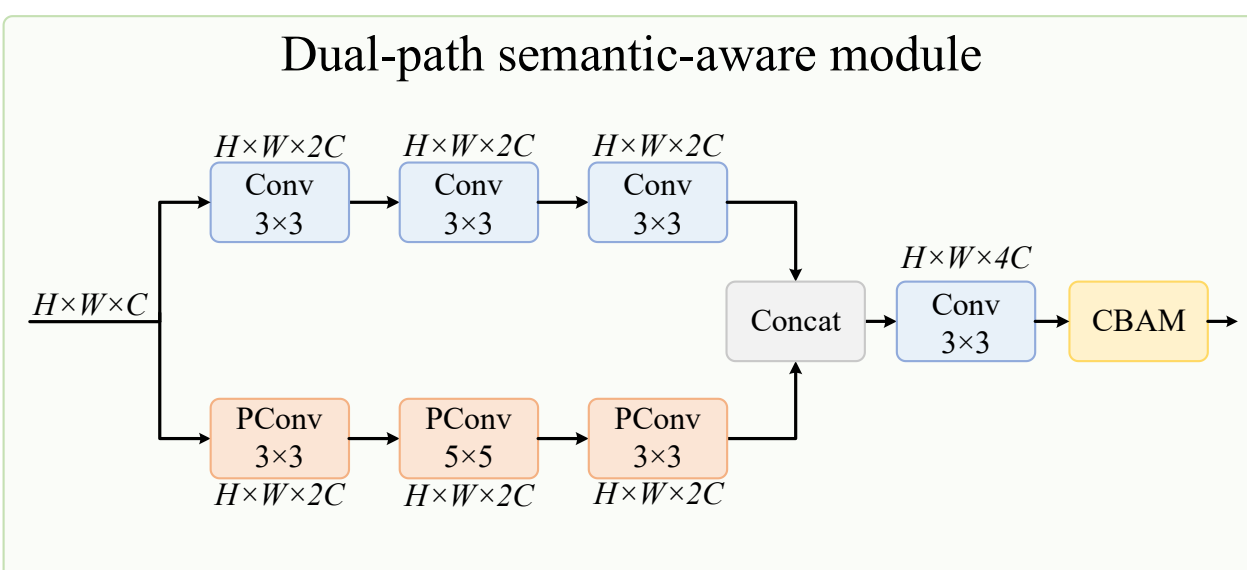

Fig. 2. Dual-path semantic perception module structure.

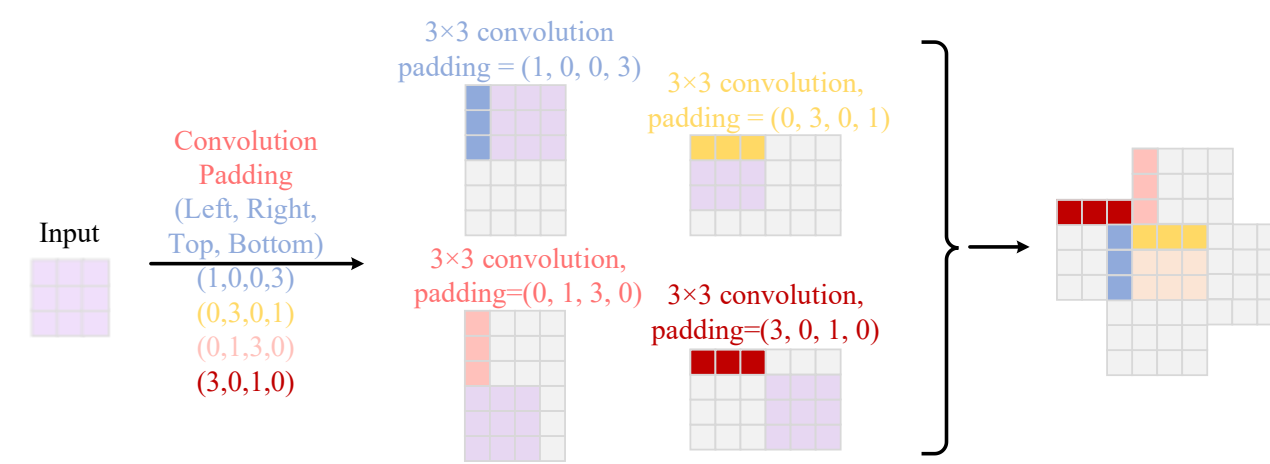

Fig. 3. Pinwheel-shaped convolution structure.

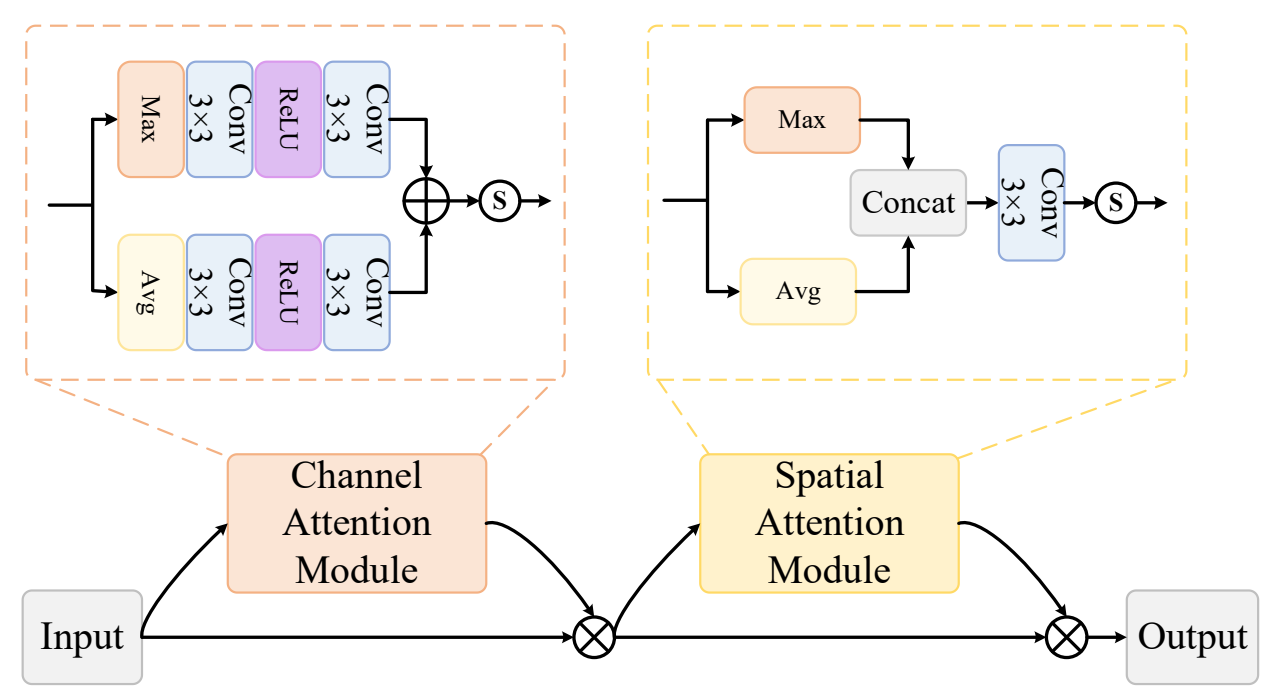

Fig. 4. Convolutional block attention module structure.

The encoder contains five encoding units, each comprising a DSM and a max-pooling operation to progressively reduce spatial resolution. DSM integrates contextual information over a broader receptive field and uses CBAM [13] to enhance target features and suppress background interference. This improves the network's sensitivity to dim, small targets.

The decoder contains four SAFMs with corresponding upsampling convolutional layers. SAFM dynamically fuses features from the encoder and decoder across scales, adaptively strengthening responses in salient regions and improving discrimination between true targets and false alarms. The upsampling convolutional layers retain the standard convolution design of U-Net and progressively restore spatial resolution.

Finally, the decoder features are passed to a detection head comprising a 3 × 3 convolution and a sigmoid activation function to generate the target prediction.

### *B. Dual-path Semantic-aware Module*

Conventional convolutional layers have fixed, limited receptive fields and cannot adequately model long-range feature dependencies. This restricts their sensitivity to dim targets, including small infrared targets. We propose DSM to improve basic feature extraction. As shown in Fig. 2, DSM consists of two parallel branches that jointly enhance feature representation. The first branch uses standard 3 × 3 convolutions to maintain local spatial consistency and preserve fine-grained details. The second branch uses pinwheel-shaped convolutions with successive kernel sizes of 3 × 3, 5 × 5, and 3 × 3 to expand the receptive field and capture broader context.

Fig. 3 shows the pinwheel-shaped convolution. Unlike conventional convolution, asymmetric padding creates direction-sensitive horizontal and vertical kernels, allowing the network to capture target structures in different directions and adapt to shape variations. Compared with other receptive-field expansion techniques, such as standard dilated convolution, pinwheel-shaped convolution requires fewer parameters and less computation. This multiscale mechanism enables DSM to model contextual semantics at different spatial scales. The features from the two branches are subsequently fused using a 3 × 3 convolution to obtain a unified representation.

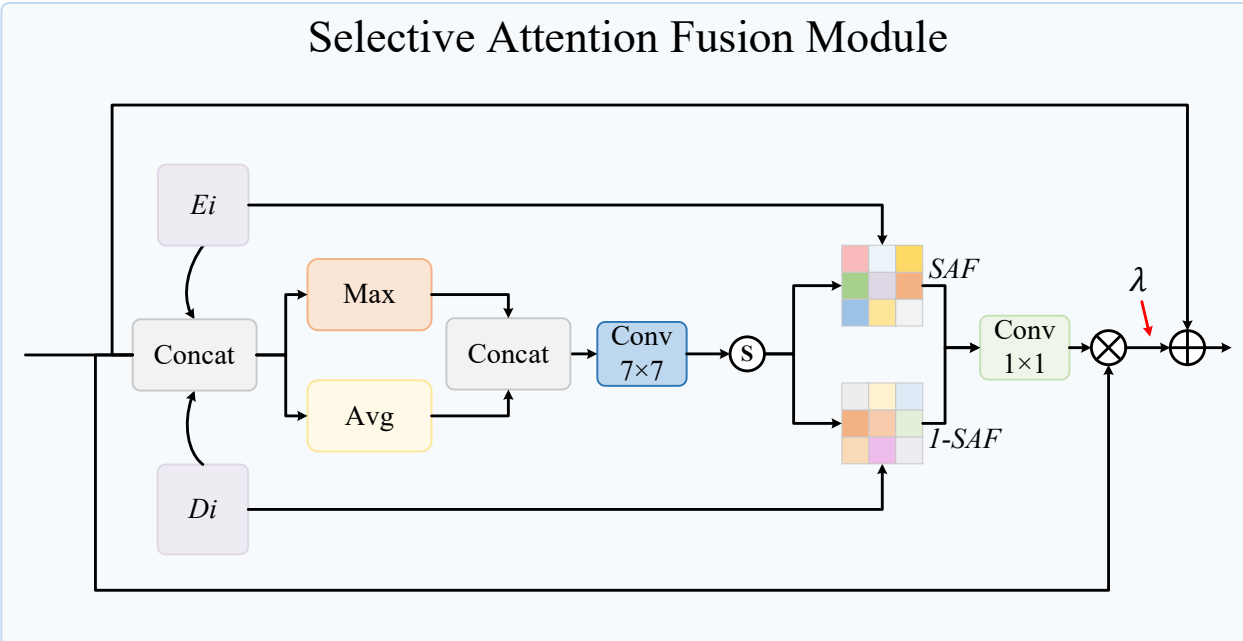

Fig. 5. Selective attention fusion module structure.

CBAM further refines the fused features. As shown in Fig. 4, CBAM cascades channel and spatial attention. Spatial attention enhances spatial feature responses, while channel attention models interchannel relationships from a global perspective and emphasizes semantic channels relevant to small target detection. Together, they improve feature discrimination.

### C. Selective Attention Fusion Module

We propose SAFM to improve discrimination between true and false targets. By introducing dynamic spatial selection into the U-Net structure, SAFM uses multiscale features to emphasize spatial regions containing informative, target-related responses, as shown in Fig. 5.

At stage i, SAFM fuses the encoder and decoder feature maps, which satisfy $E_i, D_i \in \mathbb{R}^{C_i \times \frac{H}{2^i} \times \frac{W}{2^i}}$. Here, $C_i$ is the number of channels, and H and W are the height and width of the original infrared image. The two feature maps are first concatenated along the channel dimension:

$$X_i = Concat(E_i,\, D_i), \tag{1}$$

To reduce computational complexity, channel-wise average pooling $\mathcal{P}_{avg}$ and max pooling $\mathcal{P}_{max}$ are applied separately to the fused features $X_i$:

$$SF_i^{avg} = \mathcal{P}_{avg}(X_i),\ SF_i^{max} = \mathcal{P}_{max}(X_i), \tag{2}$$

Here, $SF_i^{avg}$ and $SF_i^{max}$ are the resulting spatial feature representations. They are concatenated to form a joint spatial representation, enabling complementary spatial attention information to interact:

$$SAF_i = f_{7\times 7}\left(Concat\left(SF_i^{avg}, SF_i^{max}\right)\right), \tag{3}$$

The resulting spatial attention feature map $SAF_i$ then weights the encoder and decoder features through element-wise multiplication to obtain the fused representation $Y_i$:

$$Y_i = E_i \odot \sigma(SAF_i) + D_i \odot \sigma(1 - SAF_i), \tag{4}$$

Here, $\sigma$ denotes the sigmoid activation function, and $\odot$ denotes element-wise multiplication.

Finally, $Y_i$ is passed through a 1 × 1 convolution to expand its channel representation. The result is multiplied element-wise with the original fused features $X_i$, and a learnable scalar weight $\lambda$ is introduced to obtain the output:

$$F_{SAFM} = \lambda \odot \left(X_i \odot f_{1\times 1}(Y_i)\right) + X_i, \tag{5}$$

The parameter λ is a single global learnable scalar shared throughout SAFM and is updated during network training. It controls the fusion strength between the attention-enhanced features and the residual information, preserving the original feature structure while strengthening discriminative responses in key regions.

## IV. Experiments and Analysis

### A. Datasets and Experimental Settings

We evaluate SANet on three public benchmarks: NUAA-SIRST [5], IRSTD-1K [8], and NUDT-SIRST [7]. They contain 427, 1001, and 1327 annotated infrared images, respectively, covering diverse representative infrared scenes. The training-to-test ratios are 1:1, 4:1, and 1:1 for NUAA-SIRST, IRSTD-1K, and NUDT-SIRST, respectively. Training images are resized to 320 × 320 for NUAA-SIRST and IRSTD-1K and to 256 × 256 for NUDT-SIRST. The corresponding batch sizes are 4, 12, and 16.

The network is trained for 600 epochs using the Soft-IoU loss [37] and the AdamW optimizer. The initial learning rate is 0.001, and the weight decay is $1 \times 10^{-2}$. A cosine annealing schedule gradually reduces the learning rate to $1 \times 10^{-5}$ to support stable convergence. The segmentation threshold for the saliency maps is set to 0.5 in all experiments.

### B. Evaluation Metrics

We use four common metrics to evaluate detection performance: intersection over union (IoU), normalized intersection over union (nIoU), detection probability ($P_d$), and false alarm rate ($F_a$). They are defined as follows:

$$IoU = \frac{TP}{T + P - TP}, \tag{6}$$

$$nIoU = \frac{1}{N}\sum_{i}^{N}\frac{TP(i)}{T(i) + P(i) - TP(i)}, \tag{7}$$

$$P_d = \frac{TP}{TP + FN}, \tag{8}$$

$$F_a = \frac{FP}{FP + TN}. \tag{9}$$

Here, N is the total number of samples. T and P denote the numbers of ground-truth positive pixels and predicted positive pixels, respectively. TP, FP, and FN denote the numbers of true positive, false positive, and false negative pixels, respectively; TN denotes the number of true negative pixels.

### C. Quantitative Evaluation

We compare SANet with 14 existing IRSTD methods, comprising five traditional approaches and nine deep learning methods. The traditional methods are facet kernel and random walker (FKRW) [38], tri-layer local contrast method (TLLCM) [17], new top-hat transformation (NWMTH) [39], nonconvex optimization with an $l_p$-norm constraint (NOLC) [40], and partial sum of the tensor nuclear norm (PSTNN) [41]. The deep learning methods are asymmetric contextual modulation (ACM) [5], receptive-field and direction induced attention network (RDIAN) [42], attention-guided pyramid context network (AGPCNet) [43], dense nested attention network (DNANet) [7], U-Net in U-Net (UIUNet) [21], robust principal component analysis network (RPCANet) [44], multiscale head network (MSHNet) [45], spatial-channel cross transformer network (SCTransNet) [24], and learnable local saliency kernel network ($L^2$SKNet) [10]. The compared ACM and $L^2$SKNet models use U-Net as the backbone. For a fair comparison, all deep learning models are retrained on the same training benchmark, following the settings described in their original papers. Table I summarizes the results, from which we make the following observations.

TABLE I
COMPARISON WITH STATE-OF-THE-ARTS METHODS ON THREE DATASETS. IN IoU(%)↑, nIoU(%)↑, $P_d$(%)↑, $F_a$($10^{-6}$)↓

| Method | NUAA-SIRST | | | | IRSTD-1K | | | | NUDT-SIRST | | | |
|---|---|---|---|---|---|---|---|---|---|---|---|---|
| | IoU (%) | nIoU (%) | $P_d$ (%) | $F_a$ ($10^{-6}$) | IoU (%) | nIoU (%) | $P_d$ (%) | $F_a$ ($10^{-6}$) | IoU (%) | nIoU (%) | $P_d$ (%) | $F_a$ ($10^{-6}$) |
| FKRW | 22.06 | 28.08 | 81.77 | 16.32 | 10.39 | 16.25 | 69.54 | 24.37 | 12.67 | 21.73 | 79.51 | 67.13 |
| TLLCM | 18.66 | 27.71 | 82.24 | 16.95 | 10.26 | 18.24 | 68.55 | 24.48 | 11.08 | 23.82 | 77.97 | 67.26 |
| NWMTH | 15.77 | 17.28 | 71.12 | 55.61 | 18.94 | 16.91 | 49.26 | 21.72 | 11.72 | 10.26 | 52.71 | 46.81 |
| NOLC | 26.64 | 35.68 | 81.62 | 17.44 | 12.39 | 22.18 | 75.38 | 21.94 | 23.87 | 34.90 | 85.47 | 58.20 |
| PSTNN | 22.40 | 29.59 | 77.95 | 29.11 | 24.57 | 28.71 | 71.99 | 35.26 | 27.72 | 39.80 | 66.13 | 44.17 |
| ACM | 65.28 | 65.67 | 90.49 | 47.13 | 58.84 | 58.23 | 91.92 | 27.59 | 60.63 | 60.94 | 91.75 | 20.06 |
| RDIAN | 70.66 | 74.04 | 93.92 | 43.08 | 63.37 | 62.63 | 88.55 | 30.18 | 77.77 | 79.46 | 95.66 | 26.36 |
| AGPCNet | 75.69 | 76.60 | 96.48 | 14.99 | 66.29 | 65.23 | 92.83 | 13.12 | 88.87 | 90.64 | 97.20 | 10.02 |
| DNANet | 76.34 | 79.19 | 95.82 | 15.23 | 66.50 | 66.13 | 91.58 | 18.31 | 91.96 | 92.88 | 98.94 | 9.28 |
| UIUNet | 77.17 | 78.86 | 95.44 | 14.34 | 65.34 | 65.55 | 90.91 | 15.01 | 90.43 | 89.87 | 98.94 | 7.72 |
| RPCANet | 53.93 | 61.01 | 95.44 | 149.2 | 62.09 | 61.48 | 89.35 | 48.61 | 87.79 | 89.80 | 95.98 | 32.03 |
| MSHNet | 73.88 | 72.56 | 95.82 | 19.25 | 67.17 | 60.38 | 92.52 | 12.60 | 76.95 | 79.72 | 95.77 | 20.52 |
| SCTransNet | 75.46 | 78.06 | 95.82 | 18.80 | 66.54 | 66.36 | **93.27** | 13.68 | 93.01 | 93.50 | 98.73 | 4.80 |
| L2SKNet | 72.90 | 73.89 | 93.92 | 40.34 | 67.38 | 66.36 | 91.58 | 14.78 | 93.52 | 92.86 | 97.57 | 5.29 |
| SANet | **79.10** | **80.01** | **96.58** | **11.52** | **71.70** | **66.59** | **93.27** | **11.88** | **95.73** | **95.43** | **99.26** | **2.80** |

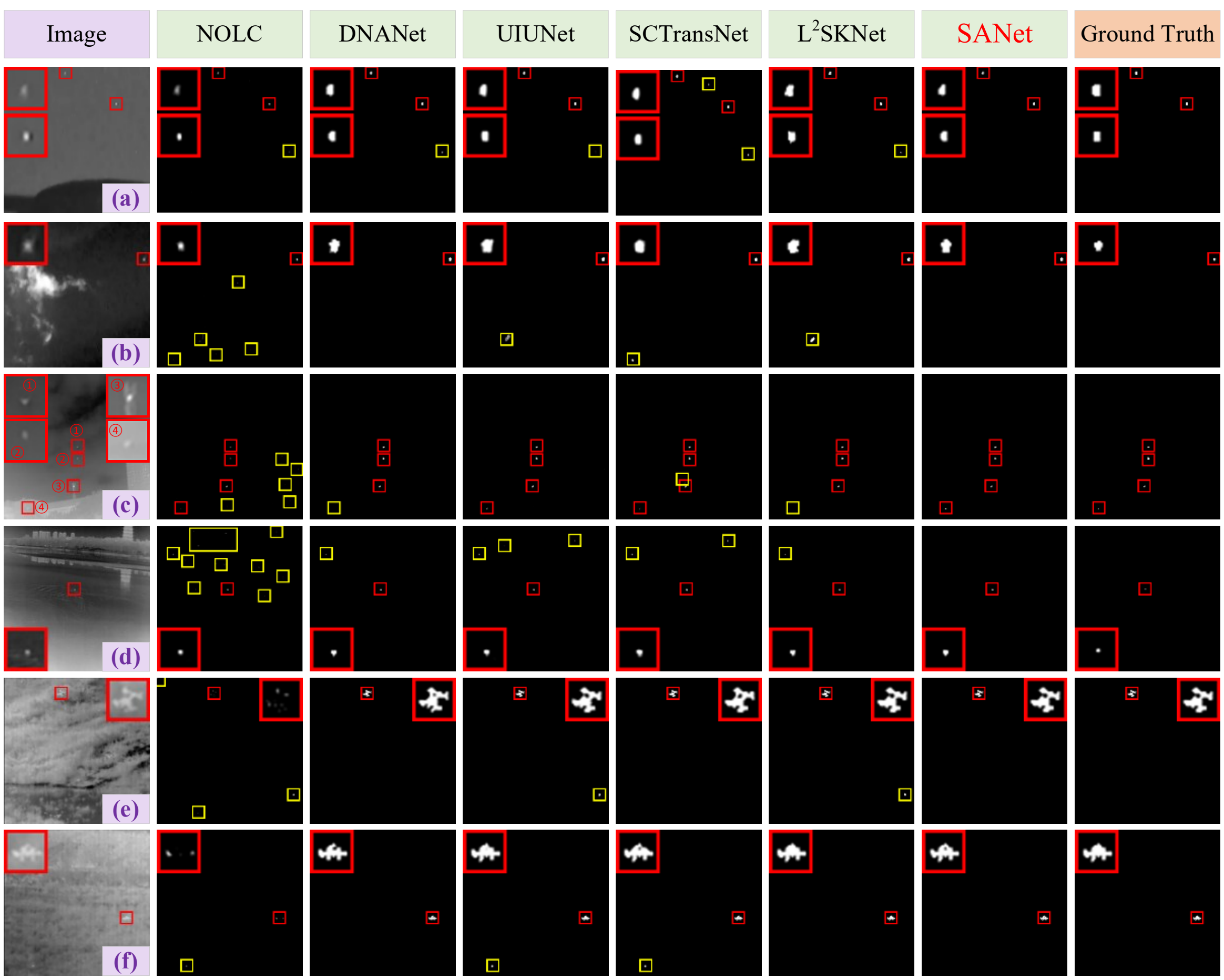

Fig. 6. Visualization of different detection methods on the NUAA-SIRST, IRSTD-1K, and NUDT-SIRST datasets. The red and yellow boxes represent correctly detected targets and false alarms, respectively.

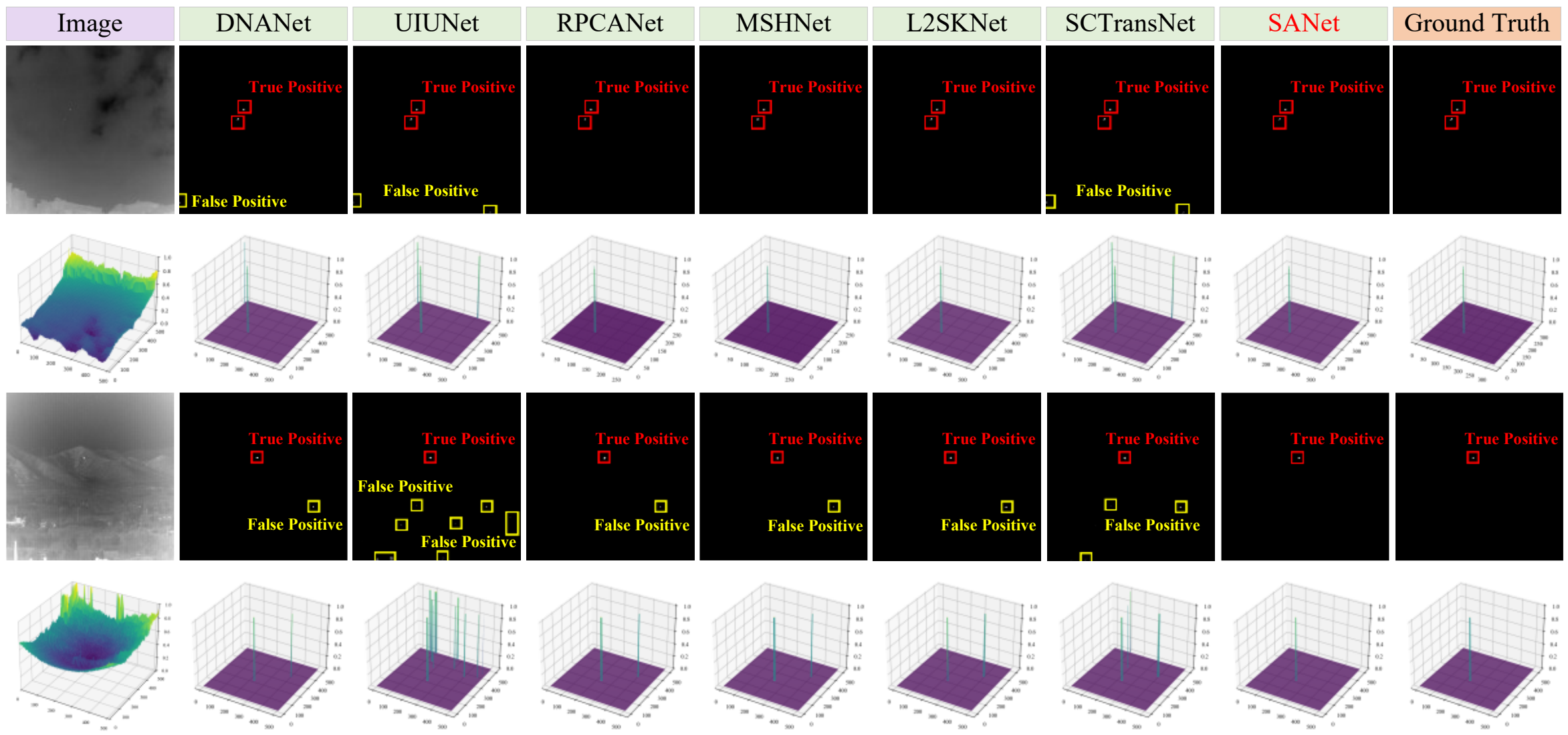

Fig. 7. 3D visualization of different IRSTD methods on the IRSTD-1K dataset. The red, blue, and yellow boxes represent correctly detected targets, missed detections, and false alarms, respectively.

TABLE II
QUANTITATIVE ABLATION EXPERIMENTS EVALUATE THE IMPACT OF THE CORE COMPONENTS DSM AND SAFM IN SANET

| Dataset | DSM | SAFM | IoU(%) | nIoU(%) | $P_d$(%) | $F_a$($10^{-6}$) |
|---|---|---|---|---|---|---|
| NUAA-SIRST | × | × | 74.91 | 79.01 | 95.44 | 28.33 |
| | ✓ | × | 78.23 | 79.51 | 95.06 | 13.86 |
| | ✓ | ✓ | **79.10** | **80.01** | **96.58** | **11.52** |
| IRSTD-1K | × | × | 64.75 | 63.24 | 91.58 | 18.77 |
| | ✓ | × | 71.46 | **67.08** | 90.57 | 16.25 |
| | ✓ | ✓ | **71.70** | 66.59 | **93.27** | **11.88** |
| NUDT-SIRST | × | × | 85.42 | 85.58 | 98.10 | 6.99 |
| | ✓ | × | 95.05 | 94.94 | 98.84 | 2.85 |
| | ✓ | ✓ | **95.73** | **95.43** | **99.26** | **2.80** |

First, SANet achieves strong performance across all three datasets. It obtains the best IoU, nIoU, and false alarm rate and the best or joint-best detection probability among the compared methods. This supports the ability of DSM and SAFM to distinguish true targets from background interference. On IRSTD-1K, its detection probability of 93.27% is equal to that of SCTransNet.

Second, deep learning methods substantially outperform traditional approaches. Handcrafted features and prior assumptions limit the ability of traditional methods to model image semantics. In contrast, deep learning methods extract semantic information from data and reduce reliance on manually tuned parameters, improving robustness to scene variations.

Third, SANet outperforms the recent L2SKNet detector on all three datasets. L2SKNet captures infrared target saliency through learnable local saliency kernels. SANet combines the proposed basic feature extraction module with selective attention fusion, producing more accurate small target detections and better overall performance.

### D. Qualitative Evaluation

Fig. 6 compares qualitative results on NUAA-SIRST in Figs. 6(a) and 6(b), IRSTD-1K in Figs. 6(c) and 6(d), and NUDT-SIRST in Figs. 6(e) and 6(f). Compared with traditional approaches, deep learning methods reduce false alarms in most cases through data-driven feature extraction, as illustrated in Figs. 6(b)–6(d). Although the other mainstream methods are competitive, they are generally less reliable than SANet in the illustrated scenes. In the complex scene in Fig. 6(c), all compared deep learning methods except UIUNet and SANet produce false alarms. However, UIUNet generates many false detections in the similarly complex scene in Fig. 6(d). SANet maintains accurate and robust detection across these scenes.

The three-dimensional visualizations in Fig. 7 further illustrate the performance of the deep learning methods. When multiple targets occur in complex backgrounds, competing methods have difficulty maintaining stable detection. UIUNet, in particular, generates many false alarms in the second scene. These results further support the robustness of SANet in complex environments.

### E. Ablation Study

#### 1) Effectiveness of the Core Components of SANet

We use conventional U-Net as the baseline and progressively introduce the two proposed modules. This stepwise integration allows us to assess the contribution of each component and the effect of combining them.

Table II shows that incorporating DSM and SAFM improves overall detection performance. Adding DSM alone increases IoU by 3.32, 6.71, and 9.63 percentage points on NUAA-SIRST, IRSTD-1K, and NUDT-SIRST, respectively. These gains indicate that DSM improves the extraction and modeling of small infrared target details during basic feature extraction.

TABLE III
QUANTITATIVE ABLATION EXPERIMENTS EVALUATE THE IMPACT OF CORE COMPONENTS IN DSM

| Dataset | Conv | PConv | CBAM | IoU(%) | nIoU(%) | $P_d$(%) | $F_a$($10^{-6}$) |
|---|---|---|---|---|---|---|---|
| NUAA-SIRST | ✓ | × | × | 74.90 | 77.53 | 93.16 | 35.74 |
| | ✓ | ✓ | × | 75.26 | 78.84 | 93.92 | 25.57 |
| | ✓ | ✓ | ✓ | **79.10** | **80.01** | **96.58** | **11.52** |
| IRSTD-1K | ✓ | × | × | 66.99 | 64.03 | 91.25 | 34.56 |
| | ✓ | ✓ | × | 67.84 | 64.41 | 93.27 | 28.45 |
| | ✓ | ✓ | ✓ | **71.70** | 66.59 | **93.27** | **11.88** |
| NUDT-SIRST | ✓ | × | × | 91.28 | 91.09 | 98.41 | 4.71 |
| | ✓ | ✓ | × | 92.95 | 92.56 | 99.05 | 6.96 |
| | ✓ | ✓ | ✓ | **95.73** | **95.43** | **99.26** | **2.80** |

TABLE IV
QUANTITATIVE ABLATION EXPERIMENTS EVALUATE THE IMPACT OF CORE COMPONENTS IN SAFM

| Dataset | Residual | $\lambda$ | IoU(%) | nIoU(%) | $P_d$(%) | $F_a$($10^{-6}$) |
|---|---|---|---|---|---|---|
| NUAA-SIRST | × | ✓ | 78.80 | **80.70** | **96.58** | 13.93 |
| | ✓ | × | 78.13 | 79.00 | 95.82 | 13.10 |
| | ✓ | ✓ | **79.10** | 80.01 | **96.58** | **11.52** |
| IRSTD-1K | × | ✓ | 70.99 | 65.58 | 89.56 | 13.49 |
| | ✓ | × | 70.04 | **67.13** | 91.92 | 26.85 |
| | ✓ | ✓ | **71.70** | 66.59 | **93.27** | **11.88** |
| NUDT-SIRST | × | ✓ | 95.23 | 94.94 | 99.05 | 2.71 |
| | ✓ | × | 95.11 | 94.72 | 98.94 | **1.54** |
| | ✓ | ✓ | **95.73** | **95.43** | **99.26** | 2.80 |

Adding SAFM further improves IoU, detection probability, and false alarm rate on all three datasets. It also improves nIoU on NUAA-SIRST and NUDT-SIRST, although nIoU on IRSTD-1K decreases from 67.08% to 66.59%. Through dynamic spatial selection, SAFM adaptively fuses multiscale features and improves discrimination between true targets and false alarms, yielding better overall detection performance.

*2) Effectiveness of the Dual-path Semantic-aware Module*

DSM improves conventional feature extraction by integrating pinwheel-shaped convolution with spatial and channel attention to generate more robust, discriminative representations. We ablate its components to assess their contributions, as shown in Table III. Pinwheel-shaped convolution improves detection by expanding the receptive field and increasing sensitivity to dim targets, particularly in terms of missed detections. However, a broader receptive field can also introduce additional background interference. For example, on NUDT-SIRST, adding pinwheel-shaped convolution increases $F_a$ from $4.71 \times 10^{-6}$ to $6.96 \times 10^{-6}$ before attention is introduced. This suggests that broad contextual information requires careful refinement to avoid redundant or noisy responses.

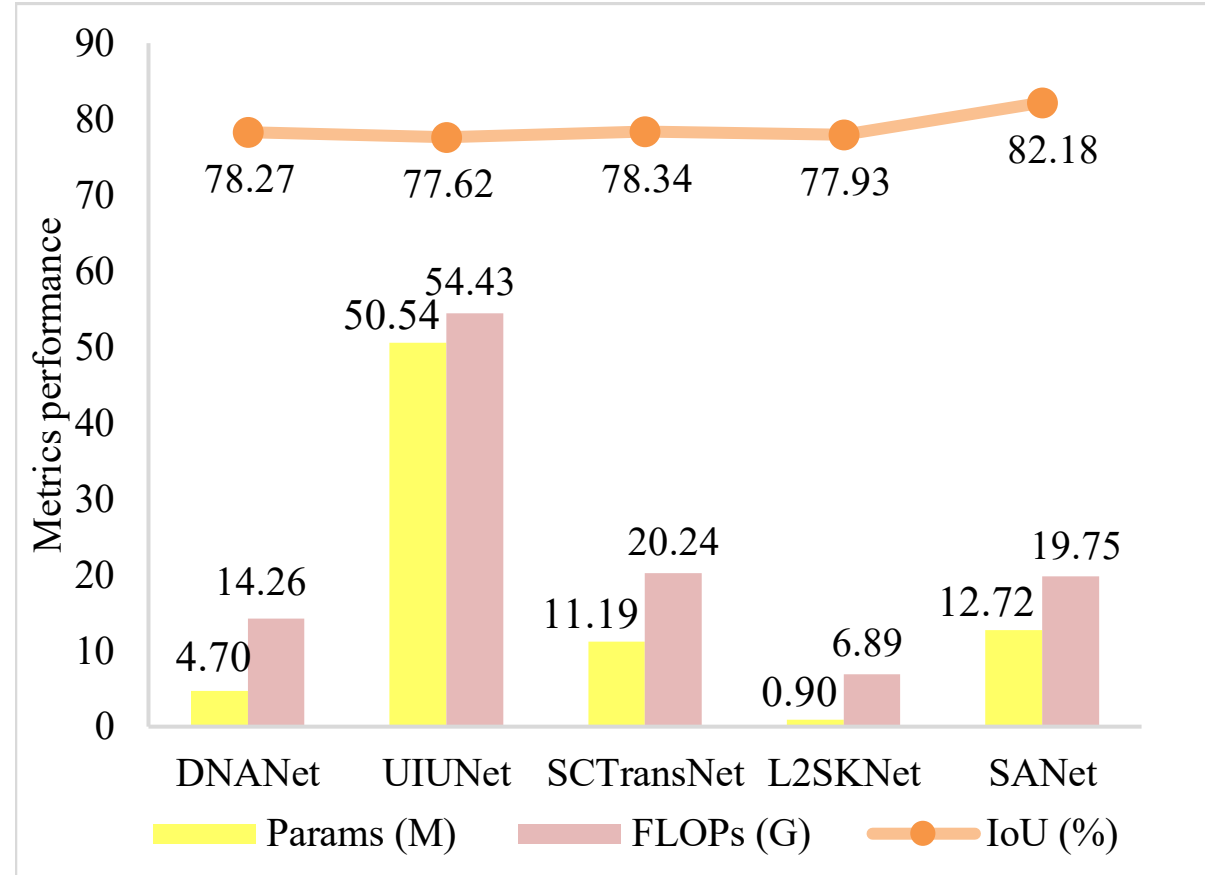

Fig.8 Performance evaluation of five algorithms across three metrics, where the IoU metric represents the average results on the three datasets.

Integrating spatial and channel attention yields larger improvements. By adaptively adjusting responses across spatial locations and channels, the attention mechanisms focus the network on potential target regions, strengthen discriminative target features, and suppress background noise.

*3) Effectiveness of the Selective Attention Fusion Module*

SAFM adaptively fuses features across scales through spatially varying weights and selectively enhances salient regions. We ablate its components to assess their individual contributions, as shown in Table IV. Removing either the residual connection or the learnable scalar λ reduces IoU. The complete model achieves IoU values of 79.10%, 71.70%, and 95.73% on the three datasets, respectively, exceeding the results obtained after removing either component. The two components therefore make complementary contributions to segmentation accuracy.

The learnable scalar improves IoU on all three datasets. On NUAA-SIRST, introducing it reduces $F_a$ from $13.10 \times 10^{-6}$ to $11.52 \times 10^{-6}$, supporting the effectiveness of selective attention fusion. The residual connection helps preserve target information and improves detection probability on multiple datasets. The improvement is particularly clear on IRSTD-1K, where $P_d$ increases from 89.56% to 93.27%. However, the full model does not minimize every metric in Table IV: some ablated variants achieve a lower false alarm rate on NUDT-SIRST or a higher nIoU on individual datasets.

## F. Model Complexity Analysis

Fig. 8 compares SANet with representative recent detectors, including DNANet, UIUNet, SCTransNet, and L2SKNet, in terms of complexity and detection performance. SANet achieves the highest mean IoU across the three benchmarks, 82.18%, exceeding the other methods by 3.84–4.56 percentage points. SANet contains 12.72 million parameters and requires 19.75 GFLOPs. Although it is not the smallest of the five models, its computational cost is moderate relative to UIUNet and SCTransNet.

These results indicate a favorable trade-off between detection accuracy and model complexity. While the architecture is somewhat more complex than the lightest alternatives, its detection performance supports its potential for practical infrared imaging applications.

## V. Conclusion

This paper proposes SANet, a selective attention network for infrared small target detection, to address dim targets, complex backgrounds, and false alarms. DSM and SAFM improve small target perception, discriminative feature representation, and background suppression.

DSM exploits the complementary properties of standard and pinwheel-shaped convolutions to expand the receptive field while preserving basic discriminative features, improving responses to small targets in complex backgrounds. SAFM replaces static skip connections in conventional U-Net with dynamic feature fusion, strengthening discrimination between true targets and false alarms.

Experiments on multiple public datasets show that SANet compares favorably with mainstream methods in detection accuracy and robustness across datasets, supporting its practical value. Future work will focus on reducing model complexity and improving detection in challenging scenes to extend its applicability to different infrared imaging tasks.